\documentclass[graybox]{svmult}

\usepackage{type1cm}        % activate if the above 3 fonts are
\usepackage{makeidx}         % allows index generation
\usepackage{graphicx}        % standard LaTeX graphics tool
\usepackage{multicol}        % used for the two-column index
\usepackage[bottom]{footmisc}% places footnotes at page bottom
\usepackage{amsfonts}
\usepackage{bm}
\usepackage{gensymb}
\usepackage{mathtools}
\usepackage{subfigure}
\usepackage{spverbatim}
\usepackage[ruled,vlined]{algorithm2e}
\usepackage{booktabs}
\usepackage{float}
\usepackage[misc]{ifsym}

\DeclarePairedDelimiter\floor{\lfloor}{\rfloor}

\usepackage{newtxtext}       %
\usepackage{newtxmath}       % selects Times Roman as basic font

\makeindex             % used for the subject index
\begin{document}

\title*{Volumetric Medical Image Segmentation: A 3D Deep Coarse-to-fine Framework and Its Adversarial Examples}
\titlerunning{Volumetric Medical Image Segmentation}
\author{Yingwei Li$^{\ast}$, Zhuotun Zhu$^{\ast}$, Yuyin Zhou, Yingda Xia, Wei Shen, Elliot K. Fishman, and Alan L. Yuille}
\authorrunning{Y. Li, Z. Zhu, Y. Zhou, \emph{et al.}}
\institute{Y. Li $\cdot$ Z. Zhu $\cdot$ Y. Zhou $\cdot$ Y. Xia $\cdot$ W. Shen $\cdot$ A. L. Yuille ${(\textrm{\Letter})}$  \at Johns Hopkins University, 3400 N Charles St, Baltimore, MD 21218 \\\email{\{yingwei.li,ztzhu,yzhou103,yxia25,wshen10,ayuille1\}@jhu.edu}
\and E. K. Fishman \at Johns Hopkins University School of Medicine, 733 N Broadway, Baltimore, MD 21205 \\\email{ efishman@jhmi.edu}}
\maketitle

\abstract*{Although deep neural networks have been a dominant method for many 2D vision tasks, it is still challenging to apply them to 3D tasks, such as medical image segmentation, due to the limited amount of annotated 3D data and limited computational resources.
In this chapter, by rethinking the strategy to apply 3D Convolutional Neural Networks to segment medical images, we propose a novel 3D-based coarse-to-fine framework to efficiently tackle these challenges. The proposed 3D-based framework outperforms their 2D counterparts by a large margin since it can leverage the rich spatial information along all three axes. We further analyze the threat of adversarial attacks on the proposed framework and show how to defense against the attack. We conduct experiments on three datasets, the NIH pancreas dataset, the JHMI pancreas dataset and the JHMI pathological cyst dataset, where the first two and the last one contain healthy and pathological pancreases respectively, and achieve the current state-of-the-art in terms of Dice-S{\o}rensen Coefficient (DSC) on all of them. Especially, on the NIH pancreas segmentation dataset, we outperform the previous best by an average of over $2\%$, and the worst case is improved by $7\%$ to reach almost $70\%$, which indicates the reliability of our framework in clinical applications.}

\abstract{Although deep neural networks have been a dominant method for many 2D vision tasks, it is still challenging to apply them to 3D tasks, such as medical image segmentation, due to the limited amount of annotated 3D data and limited computational resources.
In this chapter, by rethinking the strategy to apply 3D Convolutional Neural Networks to segment medical images, we propose a novel 3D-based coarse-to-fine framework to efficiently tackle these challenges. The proposed 3D-based framework outperforms their 2D counterparts by a large margin since it can leverage the rich spatial information along all three axes. We further analyze the threat of adversarial attacks on the proposed framework and show how to defense against the attack. We conduct experiments on three datasets, the NIH pancreas dataset, the JHMI pancreas dataset and the JHMI pathological cyst dataset, where the first two and the last one contain healthy and pathological pancreases respectively, and achieve the current state-of-the-art in terms of Dice-S{\o}rensen Coefficient (DSC) on all of them. Especially, on the NIH pancreas dataset, we outperform the previous best by an average of over $2\%$, and the worst case is improved by $7\%$ to reach almost $70\%$, which indicates the reliability of our framework in clinical applications.}

\let\thefootnote\relax\footnote{$\ast$ The first two authors contribute equally and are ordered alphabetically. The first part of this work appeared as a conference paper~\cite{ZhuXSFY18}, in which Zhuotun Zhu, Yingda Xia, and Wei Shen made contributions to. The second part was contributed by Yingwei Li, Yuyin Zhou, and Wei Shen. Elliot K. Fishman and Alan L. Yuille oversaw the entire project.}

\section{Introduction}
\label{sec:1}
Driven by the huge demands for computer-aided diagnosis systems, automatic organ segmentation from medical images, such as computed tomography (CT) and magnetic resonance imaging (MRI), has become an active research topic in both the medical image processing and computer vision communities. It is a prerequisite step for many clinical applications, such as diabetes inspection, organic cancer diagnosis, and surgical planning. Therefore, it is well worth exploring automatic segmentation systems to accelerate the computer-aided diagnosis in medical image analysis.

%Due to the insufficient number of experienced doctors and limited working efficiency, pancreatic cancers were the $7$th most common cause of cancer deaths in 2012 globally~\cite{stewart2017world}.

\begin{figure}[t]
\centering
\includegraphics[width=1\linewidth]{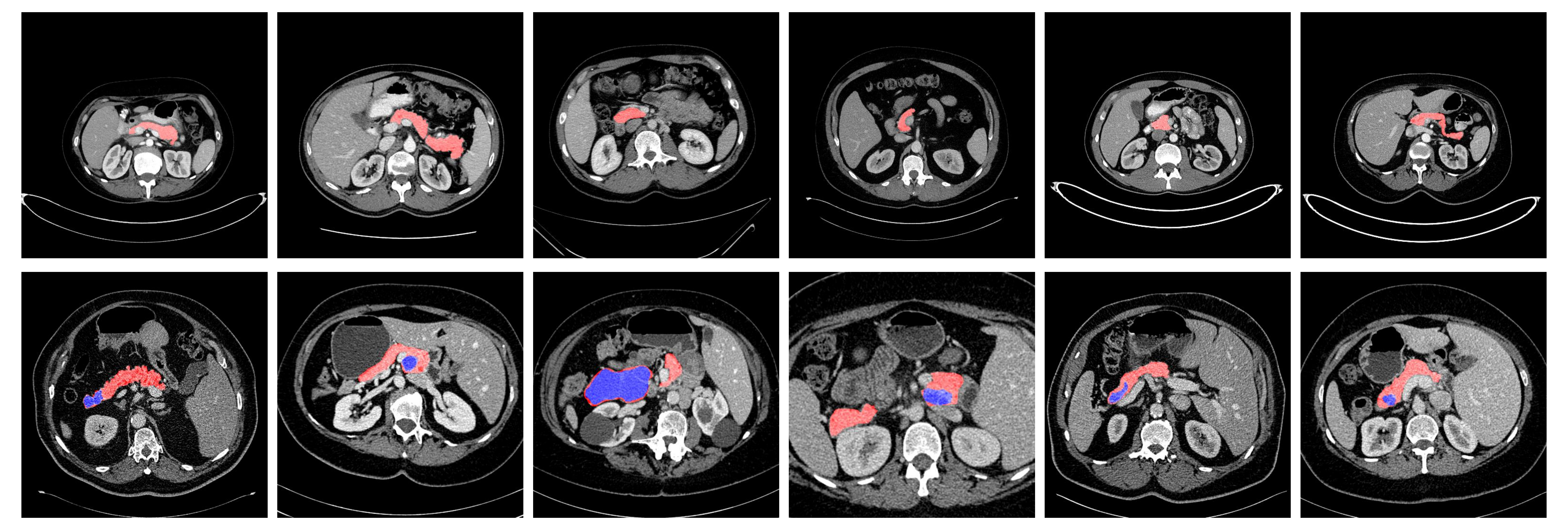}\\
   \caption{An illustration of normal pancreases on NIH dataset~\cite{roth2015deeporgan} and abnormal cystic pancreases on JHMI dataset~\cite{zhou2017deep} shown in the first and second row respectively. Normal pancreas regions are masked as red and abnormal pancreas regions are marked as blue. The pancreas usually occupies a small region in a whole CT scan. Best viewed in color.}
\label{fig:CTPancreas}
\end{figure}

In this chapter, we focus on pancreas segmentation from CT scans, one of the most challenging organ segmentation problems~\cite{zhou2017fixed}\cite{roth2015deeporgan}. As shown in Fig.~\ref{fig:CTPancreas}, the main difficulties stem from three parts: 1) the small size of the pancreas in the whole abdominal CT volume; 2) the large variations in texture, location, shape and size of the pancreas;  3) the abnormalities, like pancreatic cysts, can alter the appearance of pancreases a lot.

% 3) the significant ambiguities along the boundaries between pancreatic and non-pancreatic tissues;

% In the supplementary material, we put more examples to illustrate the three difficulties.

Following the rapid development of deep neural networks~\cite{KrizhevskySH12}\cite{SimonyanZ14a} and their successes in many computer vision tasks, such as semantic segmentation~\cite{LongSD15}\cite{ChenPKMY17}, edge detection~\cite{ShenWWBZ15}\cite{XieT15}\cite{ShenWJWY17} and 3D shape retrieval~\cite{zhu2016deep}\cite{fang20153d}, many deep learning based methods have been proposed for pancreas segmentation and have achieved considerable progress~\cite{zhou2017fixed}\cite{roth2015deeporgan}\cite{roth2016spatial}. However, these methods are based on 2D fully convolutional networks (FCNs)~\cite{LongSD15}, which perform segmentation slice by slice while CT volumes are indeed 3D data. Although these 2D methods use strategies to fuse the output from different 2D views to obtain 3D segmentation results, they inevitably lose some 3D context, which is important for capturing the discriminative features of the pancreas with respect to background regions.

An obstacle to train 3D deep segmentation networks is that it suffers from the ``out of memory" problem. 2D FCNs can accept a whole 2D slice as input, but 3D FCNs cannot be fed a whole 3D volume due to the limited GPU memory size. A common solution is to train 3D FCNs from small sub-volumes and test them in a sliding-window manner~\cite{milletari2016v}\cite{bui20173d}\cite{cciccek20163d}\cite{chen2017voxresnet}\cite{yu2017automatic}, \emph{i.e.}, performing 3D segmentation on densely and uniformly sampled sub-volumes one by one. Usually, these neighboring sampled sub-volumes overlap with each other to improve the robustness of the final 3D results. It is worth noting that the overlap size is a trade-off between the segmentation accuracy and the time cost. Setting a larger/smaller overlap size generally leads to a better/worse segmentation accuracy but takes more/less time during testing.

To address these issues, we propose a concise and effective framework to train 3D deep networks for pancreas segmentation, which can simultaneously achieve high segmentation accuracy and low time cost. Our framework is formulated into a coarse-to-fine manner. In the training stage, we first train a 3D FCN from the sub-volumes sampled from an entire CT volume. We call this {\bf\textit{ResDSN Coarse}}
model, which aims at obtaining the rough location of the target pancreas from the whole CT volume by making full use of the overall 3D context. Then, we train another 3D FCN from the sub-volumes sampled only from the ground truth bounding boxes of the target pancreas. We call this the {\bf\textit{ResDSN Fine}} model, which can refine the segmentation based on the coarse result. In the testing stage, we first apply the coarse model in the sliding-window manner to a whole CT volume to extract the most probable location of the pancreas. Since we only need a rough location for the target pancreas in this step, the overlap size is set to a small value. Afterwards, we apply the fine model in the sliding-window manner to the coarse pancreas region, but by setting a larger overlap size. Thus, we can efficiently obtain a fine segmentation result and we call the coarse-to-fine framework by {\bf\textit{ResDSN C2F}}.

Note that, the meaning of ``coarse-to-fine'' in our framework is twofold. First, it means the input region of interests (RoIs) for the {\bf\textit{ResDSN Coarse}} model and the {\bf\textit{ResDSN Fine}} model are different, \emph{i.e.}, a whole CT volume for the former one and a rough bounding box of the target pancreas for the latter one. We refer to this as coarse-to-fine RoIs, which is designed to achieve better segmentation performance. The coarse step removes a large amount of the unrelated background region, then with a relatively smaller region to be sampled as input, the fine step can much more easily learn cues which distinguish the pancreas from the local background, \emph{i.e.}, exploit local context which makes it easier to obtain a more accurate segmentation result. Second, it means the overlap sizes used for the {\bf\textit{ResDSN Coarse}} model and the {\bf\textit{ResDSN Fine}} model during inference are different, \emph{i.e.}, small and large overlap sizes for them, respectively. We refer to this as coarse-to-fine overlap sizes, which is designed for efficient 3D inference.

Recently, it is increasingly realized that deep networks are vulnerable to adversarial examples,
\emph{i.e.}, inputs that are almost indistinguishable from natural data which are imperceptible to a human, but yet classified incorrectly by the network~\cite{goodfellow2014explaining}\cite{szegedy2013intriguing}\cite{xie2017adversarial}. This problem is even more serious for medical learning systems, as they may cause incorrect decisions, which could mislead human doctors. Adversarial examples may be only a small subset of the space of all medical images, so it is possible that they will only rarely occur in real datasets. But, even so, they could potentially have major errors. Analyzing them can help medical imaging researchers to understand more about their deep network based model, with the ultimate goal of increasing robustness. In this chapter, we generate 3D adversarial examples by the gradient-based methods~\cite{goodfellow2014explaining,kurakin2016scale} and investigate the threat of these 3D adversarial examples on our framework. We also show how to defend against these adversarial examples.

The contributions of this chapter can be summarized into two aspects: (1) A novel 3D deep network based framework which leverages the rich spatial information for medical image segmentation, which achieves the state-of-the-art performance with relative low time cost on segmenting both normal and abnormal pancreases; (2) A systematic analysis about the threat of 3D adversarial examples on our framework as well as the adversarial defense methods.

The first part of this work appeared as a conference paper~\cite{ZhuXSFY18}, in which Zhuotun Zhu, Yingda Xia, and Wei Shen made contributions to. The second part was contributed by Yingwei Li, Yuyin Zhou, and Wei Shen. Elliot K. Fishman and Alan L. Yuille oversaw the entire project. This chapter extends the previous work~\cite{ZhuXSFY18} by including the analysis about the 3D adversarial attacks and defenses for our framework and more experimental results.·

% This chapter extends our previous work~\cite{ZhuXSFY18} by including the analysis about the 3D adversarial attacks and defenses for our framework and more experimental results.

%To our best knowledge, we are one of the first studies to segment the challenging {\bf\textit{normal}} and {\bf\textit{abnormal}} pancreases using 3D networks which leverage the rich spatial information. The effectiveness and efficiency of the proposed 3D coarse-to-fine framework are demonstrated on two pancreas segmentation datasets where we achieve the state-of-the-art with relative low time cost. It is worth mentioning that, although our focus is pancrease segmentation, our framework is generic and can be directly applied to segmenting other medical organs.

\begin{figure*}[t]
  \centering
  \includegraphics[width=1\linewidth]{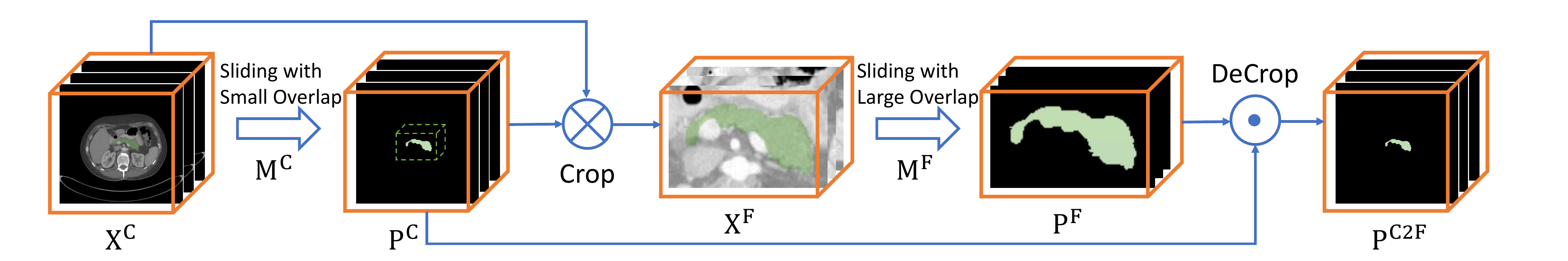}\\
   \caption{Flowchart of the proposed 3D coarse-to-fine segmentation system in the testing phase. We first apply ``ResDSN Coarse" with a small overlapped sliding window to obtain a rough pancreas region and then use the ``ResDSN Fine" model to refine the results with a large overlapped sliding window.
   Best viewed in color.}
\label{fig:Intro}
\end{figure*}

\section{Related Work}
\label{sec:2}
\subsection{Deep Learning based Medical Image Segmentation}
The medical image analysis community is facing a revolution brought by the fast development of deep networks~\cite{KrizhevskySH12}\cite{SimonyanZ14a}. Deep Convolutional Neural Networks (CNNs) based methods have dominated the research area of volumetric medical image segmentation in the last few years. Generally speaking, CNN-based methods for volumetric medical image segmentation can be divided into two major categories: 2D CNNs based and 3D CNNs based.

% 2D CNNs based methods for Volumetric Segmentation
\subsubsection{2D CNNs for Medical Image Segmentation}
2D CNNs based methods~\cite{roth2015deeporgan}\cite{roth2016spatial}\cite{HavaeiDWBCBPJL17}\cite{MoeskopsWVGLVI17}\cite{ronneberger2015u}\cite{WangZTSFY18}\cite{Yan-MultiOrgan18} performed volumetric segmentation slice by slice from different views, and then fused the 2D segmentation results to obtain a 3D Volumetric Segmentation result. In the early stage, the 2D segmentation based models were trained from image patches and tested in a patch by patch manner~\cite{roth2015deeporgan}, which is time consuming. Since the introduction of fully convolution networks (FCNs)~\cite{LongSD15}, almost all the 2D segmentation methods are built upon 2D FCNs to perform holistic slice segmentation during both training and testing. Havaei \textit{et al}~\cite{HavaeiDWBCBPJL17} proposed a two-pathway FCN architecture, which exploited both local features as well as more global contextual features simultaneously by the two pathways. Roth \textit{et al}~\cite{roth2016spatial} performed Pancreas segmentation by a holistic learning approach, which first segment pancreas regions by holistically-nested networks~\cite{XieT15} and then refine them by the boundary maps obtained by robust spatial aggregation using random forest. The U-Net~\cite{ronneberger2015u} is one of the most popular FCN architectures for medical image segmentation, which is a encoder-decoder network, but with additional short connection between encoder and decoder paths. Based on the fact that a pancreas only takes up a small fraction of the whole scan, Zhou \emph{et al.}~\cite{zhou2017fixed} proposed to find the rough pancreas region and then learn a FCN based fixed-point model to refine the pancreas region iteratively. Their method is also based on a coarse-to-fine framework, but it only considered coarse-to-fine RoIs. Besides coarse-to-fine RoIs, our coarse-to-fine method also takes coarse-to-fine overlap sizes into account, which is designed specifically for efficient 3D inference.

% 3D-based CNNs for Volumetric Segmentation
\subsubsection{3D CNNs for Medical Image Segmentation}
Although 2D CNNs based methods achieved considerable progress, they are not optimal for medical image segmentation, as they cannot make full use of the 3D context encoded in volumetric data. Several 3D CNNs based segmentation methods have been proposed. The 3D U-Net~\cite{cciccek20163d} extended the previous 2D U-Net architecture~\cite{ronneberger2015u} by replacing all 2D operations with their 3D counterparts. Based on the architecture of the 3D U-Net, the V-Net~\cite{milletari2016v} introduced residual structures~\cite{he2016deep} (short term skip connection) into each stage of the network. Chen \textit{et al}~\cite{chen2017voxresnet} proposed a deep voxel-wise residual network for 3D brain segmentation. Both I2I-3D~\cite{merkow2016dense} and 3D-DSN~\cite{dou20173d} included auxiliary supervision via side outputs into their 3D deep networks. Despite the success of 3D CNNs as a technique for segmenting the target organs, such as prostate~\cite{milletari2016v} and kidney~\cite{cciccek20163d}, very few techniques have been developed for leveraging 3D spatial information on the challenging pancreas segmentation. Gibson \emph{et al.}~\cite{gibson2018automatic} proposed the DenseVNet which is however constrained to have shallow encoders due to the computationally demanding dense connections. Roth \emph{et al.}~\cite{roth2018towards} extended 3D U-Net to segment the pancreas, while obtaining good results, this method has the following shortcomings, 1) the input of their networks is fixed to $120\times120\times120$, which is very computationally demanding due to this large volume size, 2) the rough pancreas bounding box is resampled to a fixed size as their networks input, which loses information and flexibility, and cannot deal with the intrinsic large variations of pancreas in shape and size. Therefore, we propose our 3D coarse-to-fine framework that works on both normal and abnormal CT data to ensure both low computation cost and high pancreas segmentation accuracy.
\subsection{Adversarial Attacks and Defenses for Medical Image Segmentation Networks}
Deep learning has become increasingly adopted within the medical imaging community for a wide range of tasks including classification, segmentation, detection, \emph{etc}. Though achieving tremendous success in various problems, CNNs have been demonstrated to be extremely vulnerable to adversarial examples, \emph{i.e.}, images which are crafted by human-imperceptible perturbations~\cite{goodfellow2014explaining}\cite{szegedy2013intriguing}\cite{xie2017adversarial}. Xie \emph{et al.}~\cite{xie2017adversarial} were first to make adversarial examples for semantic segmentation, which is directly related to medical image segmentation. Paschali \emph{et.al.}~\cite{paschali2018generalizability} used the code from Xie \emph{et al.}~\cite{xie2017adversarial} and showed that state-of-the-art networks such as Inception~\cite{szegedy2016rethinking} and UNet~\cite{Ronneberger_2015_UNet} are still extremely susceptible to adversarial examples for skin lesion classification and whole brain segmentation. It was also demonstrated that adversarial examples are
superior in pushing a network to its limits and evaluating its robustness in~\cite{paschali2018generalizability}. Additionally, Huang \emph{et.al.}~\cite{huang2018some} pointed out that the robustness of deep learning-based reconstruction techniques for limited angle tomography remains a concern due to its vulnerability to adversarial examples. This makes the robustness of neural networks for clinical applications an important unresolved issue. 

To alleviate such adversarial effects for clinical applications, we investigate the application of adversarial training~\cite{szegedy2013intriguing} for improving the robustness of deep learning algorithms in the medical area. Adversarial training was first proposed by Szegedy \emph{et.al.}~\cite{szegedy2013intriguing} to increase robustness by augmenting training data with adversarial examples. Madry~\emph{et.al.}~\cite{madry2017towards} further validated that adversarially trained models can be robust against white-box attacks,~\emph{i.e.}, with knowledge of the model parameters. Note that clinical applications of deep learning require a high level of safety and security~\cite{huang2018some}.
Our experiments empirically demonstrate that adversarial training can be greatly beneficial for improving the robustness of 3D deep learning-based models against adversarial examples.
\section{Method}
\subsection{A 3D Coarse-to-fine Framework for Medical Image Segmentation}\label{sec:framework}
In this section, we elaborate our 3D coarse-to-fine framework which includes a \textit{\textbf{coarse}} stage and a \textit{\textbf{fine}} stage afterwards. We first formulate a segmentation model that can be generalized to both \textit{\textbf{coarse}} stage and \textit{\textbf{fine}} stage. Later in Sec.~\ref{Sec:CoarseStage} and Sec.~\ref{Sec:FineStage}, we will customize the segmentation model to these two stages, separately.

We denote a 3D CT-scan volume by $\mathbf{X}$. This is associated with a human-labeled per-voxel annotation $\mathbf{Y}$, where both $\mathbf{X}$ and $\mathbf{Y}$ have size $W\times H\times D$, which corresponds to axial, sagittal and coronal views, separately. The ground-truth segmentation mask $\mathbf{Y}$ has a binary value $y_i$, $i = 1, \cdots, WHD$, at each spatial location $i$ where $y_i = 1$ indicates that $x_i$ is a pancreas voxel.
Denote a segmentation model by $\mathbb{M}: \mathbf{P} = {\mathbf{f}\!\left(\mathbf{X}; \boldsymbol{\Theta}\right)}$, where $\boldsymbol{\Theta}$ indicates model parameters and $\mathbf{P}$ means the binary prediction volume. Specifically in a neural network with $L$ layers and parameters $\boldsymbol{\Theta} = \{\mathcal{W}, \mathcal{B}\}$, $\mathcal{W}$ is a set of weights and $\mathcal{B}$ is a set of biases, where $\mathcal{W} = \{\mathbf{W}^1, \mathbf{W}^2, \cdots, \mathbf{W}^L\}$ and $\mathcal{B} = \{\mathbf{B}^1, \mathbf{B}^2, \cdots, \mathbf{B}^L\}$. Given that $p(y_i | x_i; \boldsymbol{\Theta})$ represents the predicted probability of a voxel $x_i$ being what is the labeled class at the final layer of the output, the negative log-likelihood loss can be formulated as:

\begin{equation}\label{Eq:SoftmaxLoss}
\mathcal{L} = \mathcal{L}(\mathbf{X}; \boldsymbol{\Theta}) = -\sum_{x_i\in \mathbf{X}}{\log(p(y_i | x_i; \boldsymbol{\Theta}))}.
\end{equation}
%If the task is binary prediction as it is for pancreas segmentation, this negative log-likelihood loss is also known as the cross entropy loss.
It is also known as the cross entropy loss in our binary segmentation setting. By thresholding $p(y_i | x_i; \boldsymbol{\Theta})$, we can obtain the binary segmentation mask $\mathbf{P}$.

We also add some auxiliary layers to the neural network, which produces side outputs under deep supervision~\cite{lee2015deeply}. These auxiliary layers form a branch network and facilitate feature learning at lower layer of the mainstream network. Each branch network shares the weights of the first $d$ layers from the mainstream network, which is denoted by $\boldsymbol{\Theta}_d = \{\mathcal{W}_d , \mathcal{B}_d\}$ and has its own weights $\widehat{\boldsymbol{\Theta}}_d$ to output the per-voxel prediction. Similarly, the loss of an auxiliary network can be formulated as:
% {\small}
\begin{equation}\label{Eq:SoftmaxLoss}
\mathcal{L}_d(\mathbf{X}; \boldsymbol{\Theta}_d, \widehat{\boldsymbol{\Theta}}_d) = \sum_{x_i\in \mathbf{X}}{-\log(p(y_i | x_i; \boldsymbol{\Theta}_d, \widehat{\boldsymbol{\Theta}}_d))},
\end{equation}
which is abbreviated as $\mathcal{L}_d$. Finally, stochastic gradient descent is applied to minimize the negative log-likelihood, which is given by following regularized objective function:
\begin{equation}\label{Eq:OverallObj}
\mathcal{L}_{overall} = \mathcal{L} + \sum_{d\in \mathcal{D}}\xi_d\mathcal{L}_d + \lambda ({\Vert \boldsymbol{\Theta} \Vert}^2 + \sum_{d\in \mathcal{D}}{\Vert \widehat{\boldsymbol{\Theta}}_d \Vert)}^2,
\end{equation}
where $\mathcal{D}$ is a set of branch networks for auxiliary supervisions, $\xi_d$ balances the importance of each auxiliary network and $l_2$ regularization is added to the objective to prevent the networks from overfitting. For notational simplicity, we keep a segmentation model that is obtained from the overall function described in Eq.~\ref{Eq:OverallObj} denoted by $\mathbb{M}: \mathbf{P} = {\mathbf{f}\!\left(\mathbf{X}; \boldsymbol{\Theta}\right)}$, where $\boldsymbol{\Theta}$ includes parameters of the mainstream network and auxiliary networks.

% \begin{equation}\label{Eq:OverallObj}
% \begin{split}
% \mathcal{L}_{overall} & = \mathcal{L}(\mathbf{X}; \boldsymbol{\Theta}) + \sum_{d\in \mathcal{D}}\xi_d\mathcal{L}_d(\mathbf{X}; \boldsymbol{\Theta}_d, \widehat{\boldsymbol{\Theta}}_d) + \\
% & \lambda ({\Vert \boldsymbol{\Theta} \Vert}^2 + \sum_{d\in \mathcal{D}}{\Vert \widehat{\boldsymbol{\Theta}}_d \Vert)}^2,
% \end{split}
% \end{equation}

\subsubsection{Coarse Stage}\label{Sec:CoarseStage}
In the \textit{\textbf{coarse}} stage, the input of ``ResDSN Coarse" is sampled from the whole CT-scan volume denoted by $\mathbf{X}^\textrm{C}$, on which the \textit{\textbf{coarse}} segmentation model $\mathbb{M}^\textrm{C}: \mathbf{P}^\textrm{C} = {\mathbf{f}^\textrm{C}\!\left(\mathbf{X}^\textrm{C}; \boldsymbol{\Theta}^\textrm{C}\right)}$ is trained on. All the $\textrm{C}$ superscripts depict the \textit{\textbf{coarse}} stage. The goal of this stage is to efficiently produce a rough binary segmentation $\mathbf{P}^\textrm{C}$ from the complex background, which can get rid of regions that are segmented as non-pancreas with a high confidence to obtain an approximate pancreas volume. Based on this approximate pancreas volume, we can crop from the original input $\mathbf{X}^\textrm{C}$ with a rectangular cube derived from $\mathbf{P}^\textrm{C}$ to obtain a smaller 3D image space $\mathbf{X}^\textrm{F}$, which is surrounded by simplified and less variable context compared with $\mathbf{X}^\textrm{C}$. The mathematic definition of $\mathbf{X}^\textrm{F}$ is formulated as:
\begin{equation}\label{Eq:CoarseSegmentation}
\mathbf{X}^\textrm{F} = \textrm{Crop}[\mathbf{X}^\textrm{C}\otimes\mathbf{P}^\textrm{C}; \mathbf{P}^\textrm{C}, m],
\end{equation}
where $\otimes$ means an element-wise product. The function $\textrm{Crop}[\mathbf{X};\mathbf{P}, m]$ denotes cropping $\mathbf{X}$ via a rectangular cube that covers all the $1$'s voxels of a binary volume $\mathbf{P}$ added by a padding margin $m$ along three axes. Given $\mathbf{P}$, the functional constraint imposed on $\mathbf{X}$ is that they have exactly the same dimensionality in 3D space. The padding parameter $m$ is empirically determined in experiments, where it is used to better segment the boundary voxels of pancreas during the \textit{\textbf{fine}} stage. The $\textrm{Crop}$ operation acts as a dimensionality reduction to facilitate the fine segmentation, which is crucial to cut down the consuming time of segmentation. It is well-worth noting that the 3D locations of the rectangular cube which specifies where to crop $\mathbf{X}^\textrm{F}$ from $\mathbf{X}^\textrm{C}$ is recorded to map the \textit{\textbf{fine}} segmentation results back their positions in the full CT scan.

\subsubsection{Fine Stage}\label{Sec:FineStage}
In the \textit{\textbf{fine}} stage, the input of the ConvNet is sampled from the cropped volume $\mathbf{X}^\textrm{F}$, on which we train the \textit{\textbf{fine}} segmentation model $\mathbb{M}^\textrm{F}: \mathbf{P}^\textrm{F} = {\mathbf{f}^\textrm{F}\!\left(\mathbf{X}^\textrm{F}; \boldsymbol{\Theta}^\textrm{F}\right)}$, where the $\textrm{F}$ superscripts indicate the \textit{\textbf{fine}} stage. The goal of this stage is to refine the coarse segmentation results from previous stage. In practice, $\mathbf{P}^\textrm{F}$ has the same volumetric size of $\mathbf{X}^\textrm{F}$, which is smaller than the original size of $\mathbf{X}^\textrm{C}$.

\subsubsection{Coarse-to-Fine Segmentation}\label{Sec:Coarse2Fine}
Our segmentation task is to give a volumetric prediction on every voxel of $\mathbf{X}^\textrm{C}$, so we need to map the $\mathbf{P}^\textrm{F}$ back to exactly the same size of $\mathbf{X}^\textrm{C}$ given by:
\begin{equation}\label{Eq:C2FSegmentation}
\mathbf{P}^\textrm{C2F} = \textrm{DeCrop}[\mathbf{P}^\textrm{F}\odot\mathbf{P}^\textrm{C}; \mathbf{X}^\textrm{F}, \mathbf{X}^\textrm{C}],
\end{equation}
where $\mathbf{P}^\textrm{C2F}$ denotes the final volumetric segmentation, and $\odot$ means an element-wise replacement, and $\textrm{DeCrop}$ operation defined on $\mathbf{P}^\textrm{F}, \mathbf{P}^\textrm{C}, \mathbf{X}^\textrm{F} \textrm{ and } \mathbf{X}^\textrm{C}$ is to replace a pre-defined rectangular cube inside $\mathbf{P}^\textrm{C}$ by $\mathbf{P}^\textrm{F}$, where the replacement locations are given by the definition of cropping $\mathbf{X}^\textrm{F}$ from $\mathbf{X}^\textrm{C}$ given in Eq.~\ref{Eq:CoarseSegmentation}.

All in all, our entire 3D-based coarse-to-fine segmentation framework during testing is illustrated in Fig~\ref{fig:Intro}.

\begin{figure*}
  \centering
  \includegraphics[width=1\linewidth]{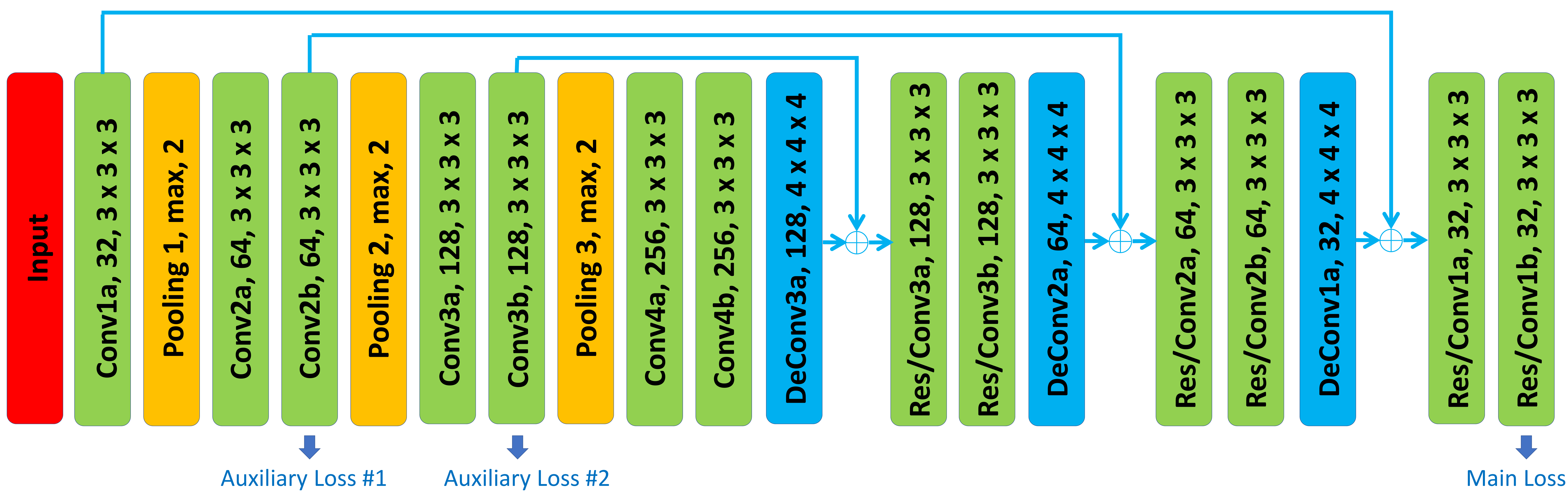}\\
  \caption{Illustration of our 3D convolutional neural network for volumetric segmentation. The encoder path is composed from ``Conv1a" to ``Conv4b" while the decoder path is from ``DeConv3a" to ``Res/Conv1b". Each convolution or deconvolution layer consists of one convolution followed by a BatchNorm and a ReLU. To clarify, ``Conv1a, 32, $3\times 3 \times 3$" means the convolution operation with $32$ channels and a kernel size of $3\times 3 \times 3$. ``Pooling 1, max, $2$" means the max pooling operation with kernel size of $2\times 2 \times 2$ and a stride of two. Long residual connections are illustrated by the blue concrete lines. Blocks with same color mean the same operations. Best viewed in color.}\label{fig:ResDSN}
%\vspace{-0.5cm}
% Blue concrete line with arrow in the end denotes the long-residual connection. Red dashed lines circle the encoder path while blue dashed lines circle the decoder path.
\end{figure*}

\subsubsection{Network Architecture}\label{Sec:NetworkArchietcture}
As shown in Fig.~\ref{fig:ResDSN}, we provide an illustration of our convolutional network architecture. Inspired by V-Net~\cite{milletari2016v}, 3D U-Net~\cite{cciccek20163d}, and VoxResNet~\cite{chen2017voxresnet}, we have an encoder path followed by a decoder path each with four resolution steps. The left part of network acts as a feature extractor to learn higher and higher level of representations while the right part of network decompresses compact features into finer and finer resolution to predict the per-voxel segmentation. The padding and stride of each layer (Conv, Pooling, DeConv) are carefully designed to make sure the densely predicted output is the same size as the input.

The encoder sub-network on the left is divided into different steps that work on different resolutions. Each step consists of one to two convolutions, where each convolution is composed of $3\times 3\times 3$ convolution followed by a batch normalization (BN~\cite{ioffe2015batch}) and a rectified linear unit (ReLU~\cite{nair2010rectified}) to reach better convergence, and then a max pooling layer with a kernel size of $2\times 2\times 2$ and strides of two to reduce resolutions and learn more compact features. The downsampling operation implemented by max-pooling can reduce the size of the intermediate feature maps while increasing the size of the receptive fields. Having fewer size of activations makes it possible to double the number of channels during feature aggregation given the limited computational resource.
% than one of the previous layer

The decoder sub-network on the right is composed of several steps that operate on different resolutions as well. Each step has two convolutions with each one followed by a BatchNorm and a ReLU, and afterwards a Deconvolution with a kernel size of $4\times 4 \times 4$ and strides of two is connected to expand the feature maps and finally predict the segmentation mask at the last layer. The upsampling operation that is carried out by deconvolution enlarges the resolution between each step, which increases the size of the intermediate activations so that we need to halve the number of channels due to the limited memory of the GPU card.

Apart from the left and right sub-networks, we impose a residual connection~\cite{he2016deep} to bridge short-cut connections of features between low-level layers and high-level layers. During the forward phase, the low-level cues extracted by networks are directly added to the high-level cues, which can help elaborate the fine-scaled segmentation, \emph{e.g.}, small parts close to the boundary which may be ignored during the feature aggregation due to the large size of receptive field at high-level layers.
As for the backward phase, the supervision cues at high-level layers can be back-propagated through the short-cut way via the residual connections. This type of mechanism can prevent networks from gradient vanishing and exploding~\cite{glorot2010understanding}, which hampers network convergence during training.

We have one mainstream loss layer connected from ``Res/Conv1b" and another two auxiliary loss layers connected from ``Conv2b" and ``Conv3b" to the ground truth label, respectively. For the mainstream loss in ``Res/Conv1b" at the last layer which has the same size of data flow as one of the input, a $1 \times 1 \times 1$ convolution is followed to reduce the number of channels to the number of label classes which is $2$ in our case. As for the two auxiliary loss layers, deconvolution layers are connected to upsample feature maps to be the same as input.

The deep supervision imposed by auxiliary losses provides robustness to hyper-parameters choice, in that the low-level layers are guided by the direct segmentation loss, leading to faster convergence rate. Throughout this work, we have two auxiliary branches where the default parameters are $\xi_1 = 0.2$ and $\xi_2 = 0.4$ in Eq.~\ref{Eq:OverallObj} to control the importance of deep supervisions compared with the major supervision from the mainstream loss for all segmentation networks.

\begin{table}[htb]
\footnotesize
\begin{center}
\begin{tabular}{lcccc}\toprule
Method 							& Long Res 	& Short Res & Deep Super & Loss\\
\hline
ResDSN (Ours) 							& Sum		& No		& Yes 		&CE \\
\hline
FResDSN 						& Sum		& Sum		& Yes 		&CE \\
SResDSN 						& No		& Sum		& Yes 		&CE \\
\hline
3D U-Net~\cite{cciccek20163d}	& Concat	& No		& No		&CE\\
V-Net~\cite{milletari2016v} 	& Concat	& Sum		& No		&DSC\\
VoxResNet~\cite{chen2017voxresnet}	& No 	& Sum		& Yes		&CE\\
MixedResNet~\cite{yu2017volumetric}	& Sum	& Sum 		& Yes		&CE\\
3D DSN~\cite{dou20173d} 		& No		& No		& Yes 		&CE \\
3D HED~\cite{merkow2016dense} 	& Concat	& No		& Yes 		&CE \\
\bottomrule
\end{tabular}
\end{center}
\caption{
    Configurations comparison of different 3D segmentation networks on medical image analysis.
    For all the abbreviated phrases, ``Long Res" means long residual connection, ``Short Res" means short residual connection, ``Deep Super" means deep supervision implemented by auxiliary loss layers, ``Concat" means concatenation, ``DSC" means Dice-S{\o}rensen Coefficient and ``CE" means cross-entropy. For residual connection, it has two types: concatenation (``Concat") or element-wise sum (``Sum").
}
\label{Tab:Config3DSegmentation}
\end{table}

% Correct one corresponding problem after the submission. :(

As shown in Table~\ref{Tab:Config3DSegmentation}, we give the detailed comparisons of network configurations in terms of four aspects: long residual connection, short residual connection, deep supervision and loss function. Our backbone network architecture, named as ``ResDSN", is proposed with different strategies in terms of combinations of long residual connection and short residual connection compared with VoxResNet~\cite{chen2017voxresnet}, 3D HED~\cite{merkow2016dense}, 3D DSN~\cite{dou20173d} and MixedResNet~\cite{yu2017volumetric}. In this table, we also depict ``FResDSN" and ``SResDSN", where ``FResDSN" and ``SResDSN" are similar to MixedResNet~\cite{yu2017volumetric} and VoxResNet~\cite{chen2017voxresnet}, respectively. As confirmed by our quantitative experiments in Sec.~\ref{sec:ablation}, instead of adding short residual connections to the network, \emph{e.g.}, ``FResDSN" and ``SResDSN", we only choose the long residual element-wise sum, which can be more computationally efficient while even performing better than the ``FResDSN" architecture which is equipped with both long and short residual connections. Moreover, ResDSN has noticeable differences with respect to the V-Net~\cite{milletari2016v} and 3D U-Net~\cite{cciccek20163d}. On the one hand, compared with 3D U-Net and V-Net which concatenate the lower-level local features to higher-level global features, we adopt the element-wise sum between these features, which outputs less number of channels for efficient computation. On the other hand, we introduce deep supervision via auxiliary losses into the network to yield better convergence.

\subsection{3D Adversarial Examples}
\label{sec:attack}
In this section, we discuss how to generate 3D adversarial examples for our segmentation framework as well as the defense method. 
We follow the notations defined in Sec.~\ref{sec:framework}, \emph{i.e.}, $\mathbf{X}$ denotes a 3D CT-scan volume, $\mathbf{Y}^{\text{true}}$ denotes the corresponding ground-truth label, and $\mathcal{L}(\mathbf{X}; \bm{\Theta})$ denotes the network loss function. To generate the adversarial example, the goal is to maximize the loss $\mathcal{L}(\mathbf{X} + \mathbf{r}; \bm{\Theta})$ for the image $\mathbf{X}$, under the constraint that the generated adversarial example $\mathbf{X}^{\text{adv}}=\mathbf{X}+\mathbf{r}$ should look visually similar to the original image $\mathbf{X}$ and the corresponding predicted label $\mathbf{Y}^{\text{adv}}\neq \mathbf{Y}^{\text{true}}$. By imposing additional constraints such as~$||\mathbf{r}||_{\infty}\leq\epsilon$, we can restrict the perturbation to be small enough to be imperceptible to humans.

\subsubsection{Attack Methods}
\label{sec:AttackMethods}
As for 3D adversarial attacking, we mainly adopt the gradient-based methods. They are
\begin{itemize}
\item \textbf{Fast Gradient Sign Method (FGSM):} FGSM~\cite{goodfellow2014explaining} is the first member in this attack family, which finds the adversarial perturbations in the direction of the loss gradient $\nabla_{\mathbf{X}} \mathcal{L}(\mathbf{X}; \bm{\Theta})$. The update equation is 
\begin{equation}
\label{eq:FGSM}
\mathbf{X}^{\text{adv}} = \mathbf{X} + \epsilon\cdot\text{sign}( \nabla_{\mathbf{X}} \mathcal{L}(\mathbf{X}; \bm{\Theta})).
\end{equation}

\item \textbf{Iterative Fast Gradient Sign Method (I-FGSM):} An extended iterative version of FGSM~\cite{kurakin2016scale}, which can be expressed as
\begin{align} \label{eq:IFGSM update}
\small
&\mathbf{X}_{0}^{\text{adv}} = \mathbf{X}   \\ 
&\mathbf{X}_{n+1}^{\text{adv}} = \text{Clip}_{\mathbf{X}}^{\epsilon} \left \{\mathbf{X}_{n}^{\text{adv}} + \alpha\cdot\text{sign}(\nabla_{\mathbf{X}} \mathcal{L}(\mathbf{X}_{n}^{adv}; \bm{\Theta})) \right \}, \nonumber 
\end{align}
where $\text{Clip}_{\mathbf{X}}^{\epsilon}$ indicates the resulting images are clipped within the $\epsilon$-ball of the original image $\mathbf{X}$, $n$ is the iteration number and $\alpha$ is the step size.
\end{itemize}

\subsubsection{Defending against 3D adversarial examples}
\label{sec: defense}
Following~\cite{madry2017towards}, defending against adversarial examples can be expressed as a saddle point problem, which comprises of an inner maximization problem and an outer minimization problem. More precisely, our objective for defending against 3D adversarial examples is formulated as following:

\begin{equation}
\label{eq:minmax}
	\min_{\mathbf{\Theta}} \rho(\bm{\Theta}),\quad \text{ where }\quad \rho(\bm{\Theta}) =
    \mathbb{E}_{(\mathbf{X})\sim\mathcal{D}}\left[\max_{\mathbf{r}\in \mathcal{S}}
    \mathcal{L}(\mathbf{X}+\mathbf{r};\bm{\Theta})\right].
\end{equation}
$\mathcal{S}$ and $\mathcal{D}$ denote the set of allowed perturbations and the data distribution, respectively.
%\textcolor{red}{Give experiments that why we don't use the DICE loss.}
%V-Net~\cite{milletari2016v} introduces a new objective function, which aims to directly optimize the Dice coefficient formulated as: ${2\sum_i^{WHD}p_iy_i}/{(\sum_i^{WHD}p_i^2 + \sum_i^{WHD}y_i^2)}$ given the ground truth $y_i$ and the corresponding binary prediction $p_i$ across every spatial position $i$. The main reason that we do not use Dice loss for training is that it only has gradients on the pancreas positions, which throws away patches without any pancreas voxels, \emph{i.e.}, negative patches, leading to many false positive predictions if we are using a sliding window for both training and testing phases.

\section{Experiments}
In this section, we demonstrate our experimental results, which consists of two parts. In the first part, we show the performance of our framework on pancreas segmentation. We first describe in detail how we conduct training and testing on the \textit{\textbf{coarse}} and \textit{\textbf{fine}} stages, respectively. Then we give the comparison results on three pancreas datasets: the NIH pancreas dataset~\cite{roth2015deeporgan}, the JHMI pathological cyst dataset~\cite{zhou2017deep} and the JHMI pancreas dataset. In the second part, we discuss the adversarial attack and defense results on our framework. 
%In this section, we first describe in detail how we conduct training and testing in the \textit{\textbf{coarse}} and \textit{\textbf{fine}} stages, respectively. Then we are going to compare our proposed method with previous state-of-the-art on two pancreas datasets: NIH pancreas dataset~\cite{roth2015deeporgan} and JHMI pathological pancreas dataset~\cite{zhou2017deep}.
\subsection{Pancreas Segmentation}
\subsubsection{Network Training and Testing}\label{Sec:NetworkTrTs}
%\textcolor{red}{Elaborate details on the training and testing of coarse and fine stages.}
All our experiments were run on a desktop equipped with the NVIDIA TITAN X (Pascal) GPU and deep neural networks were implemented based on the CAFFE~\cite{jia2014caffe} platform customized to support 3D operations for all necessary layers, \emph{e.g.}, ``convolution", ``deconvolution" and ``pooling", \emph{etc}. For the data pre-processing, we simply truncated the raw intensity values to be in $[-100, 240]$ and then normalized each raw CT case to have zero mean and unit variance to decrease the data variance caused by the physical processes~\cite{gravel2004method} of medical images. As for the data augmentation in the training phase, unlike sophisticated processing used by others, \emph{e.g.}, elastic deformation~\cite{milletari2016v}\cite{ronneberger2015u}, we utilized simple but effective augmentations on all training patches, \emph{i.e.}, rotation ($90\degree, 180\degree, \textrm{ and } 270\degree$) and flip in all three axes (axial, sagittal and coronal), to increase the number of 3D training samples which can alleviate the scarce of CT scans with expensive human annotations. Note that different CT cases have different physical resolutions, but we keep their resolutions unchanged. The input size of all our networks is denoted by $W_I\times H_I\times D_I$, where $ W_I = H_I = D_I =64$.

For the \textit{\textbf{coarse}} stage, we randomly sampled $64 \times 64 \times 64$ sub-volumes from the whole CT scan in the training phase. In this case, a sub-volume can either cover a portion of pancreas voxels or be cropped from regions with non-pancreas voxels at all, which acts as a hard negative mining to reduce the false positive. In the testing phase, a sliding window was carried out to the whole CT volume with a \textit{\textbf{coarse}} stepsize that has small overlaps within each neighboring sub-volume. Specifically, for a testing volume with a size of $W \times H \times D$, we have a total number of $(\floor{\frac{W}{W_I}} + n) \times (\floor{\frac{H}{H_I}} + n) \times (\floor{\frac{D}{D_I}} + n)$ sub-volumes to be fed into the network and then combined to obtain the final prediction, where $n$ is a parameter to control the sliding overlaps that a larger $n$ results in a larger overlap and vice versa. In the \textit{\textbf{coarse}} stage for the low time cost concern, we set $n = 6$ to efficiently locate the rough region of pancreas $\mathbf{X}^\textrm{F}$ defined in Eq.~\ref{Eq:CoarseSegmentation} from the whole CT scan $\mathbf{X}^\textrm{C}$.

For the \textit{\textbf{fine}} stage, we randomly cropped $64 \times 64 \times 64$ sub-volumes constrained to be from the pancreas regions defined by ground-truth labels during training. In this case, a training sub-volume was assured to cover pancreatic voxels, which was specifically designed to be capable of segmentation refinement. In the testing phase, we only applied the sliding window on $\mathbf{X}^\textrm{F}$ with a size of $W_F\times H_F\times D_F$. The total number of sub-volumes to be tested is $(\floor{\frac{W_F}{W_I}} + n)\times (\floor{\frac{H_F}{H_I}} + n)\times (\floor{\frac{D_F}{D_I}} + n)$. In the \textit{\textbf{fine}} stage for the high accuracy performance concern, we set $n = 12$ to accurately estimate the pancreatic mask $\mathbf{P}^\textrm{F}$ from the rough segmentation volume $\mathbf{X}^\textrm{F}$. In the end, we mapped the $\mathbf{P}^\textrm{F}$ back to $\mathbf{P}^\textrm{C}$ to obtain $\mathbf{P}^\textrm{C2F}$ for the final pancreas segmentation as given in Eq.~\ref{Eq:C2FSegmentation}, where the mapping location is given by the cropped location of $\mathbf{X}^\textrm{F}$ from $\mathbf{X}^\textrm{C}$.

After we get the final binary segmentation mask, we denote $\mathcal{P}$ and $\mathcal{Y}$ to be the set of pancreas voxels in the prediction and ground truth, separately, \emph{i.e.}, $\mathcal{P} = \{i | p_i = 1\}$ and $\mathcal{Y} = \{i | y_i = 1\}$. The evaluation metric is defined by the Dice-S{\o}rensen Coefficient (DSC) formulated as $\text{DSC}(\mathcal{P}, \mathcal{Y}) = \frac{2\times |\mathcal{P}\cap \mathcal{Y}|}{|\mathcal{P}| + |\mathcal{Y}|}$. This evaluation measurement ranges in $[0, 1]$ where $1$ means a perfect prediction.

\subsubsection{NIH Pancreas Dataset}
\label{sec:NIHDataset}
% a threshold of
We conduct experiments on the NIH pancreas segmentation dataset~\cite{roth2015deeporgan}, which contains $82$ contrast-enhanced abdominal CT volumes provided by an experienced radiologist. The size of CT volumes is $512\times512\times D$, where $D\in [181, 466]$ and their spatial resolutions are $w\times h\times d$, where $d = 1.0\textrm{mm}$ and $w = h$ that ranges from $0.5\textrm{mm}$ to $1.0\textrm{mm}$. Data pre-processing and data augmentation were described in Sec.~\ref{Sec:NetworkTrTs}. Note that we did not normalize the spatial resolution into the same one since we wanted to impose the networks to learn to deal with the variations between different volumetric cases. Following the training protocol~\cite{roth2015deeporgan}, we perform $4$-fold cross-validation in a random split from $82$ patients for training and testing folds, where each testing fold has $21, 21, 20$ and $20$ cases, respectively. We trained networks illustrated in Fig.~\ref{fig:ResDSN} by SGD optimizer with a $16$ mini-batch, a $0.9$ momentum, a base learning rate to be $0.01$ via polynomial decay (the power is $0.9$) in a total of $80\rm{,}000$ iterations, and the weight decay $0.0005$. Both training networks in the \textit{\textbf{coarse}} and \textit{\textbf{fine}} stages shared the same training parameter settings except that they took a $64 \times 64 \times 64$ input sampled from different underlying distributions described in Sec.~\ref{Sec:NetworkTrTs}, which included the details of testing settings as well. We average the score map of overlapped regions from the sliding window and throw away small isolated predictions whose portions are smaller than $0.2$ of the total prediction, which can remove small false positives. For DSC evaluation, we report the average with standard deviation, max and min statistics over all $82$ testing cases as shown in Table~\ref{Tab:NIHC2FSegmentation}.

\begin{table}[tb]
\footnotesize
\begin{center}
\begin{tabular}{lccc}\toprule
Method           & {Mean DSC}       		& {Max DSC}    		& {Min DSC} \\
\hline
% {ResDSN C2C (Ours)} & $\bm{84.59\pm{5.35}}\%$ 				& $\bm{91.25}\%$       	&$\bm{65.80}\%$ \\
{ResDSN C2F (Ours)}  		& $\bm{84.59\pm{4.86}}\%$ 				& $\bm{91.45}\%$       	&$\bm{69.62}\%$ \\
{ResDSN Coarse (Ours)}  		& $83.18\pm{6.02}\%$ 				& $91.33\%$       	&$58.44\%$ \\
\hline
{Cai \emph{et al.}~\cite{cai2017improving}}   & $82.4\pm{6.7}\%$          & $90.1\%$          & $60.0\%$          \\
{Zhou \emph{et al.}~\cite{zhou2017fixed}} 	& $82.37\pm{5.68}\%$ 				&$90.85\%$ 			& $62.43\%$\\
{Dou\footnotemark[1] \emph{et al.}~\cite{dou20173d}} &$82.25\pm{5.91}\%$ 				&$90.32\%$			&$62.53\%$ \\
{Roth \emph{et al.}~\cite{roth2016spatial}}		&$78.01\pm{8.20}\%$ 				&$88.65\%$			&$34.11\%$ \\
{Yu\footnotemark[1] \emph{et al.}~\cite{yu2017automatic}} &$71.96\pm{15.34}\%$ 				&$89.27\%$			&$0\%$ \\
%{Roth \emph{et al.}~\cite{roth2015deeporgan}}		&$71.42\pm{10.11}\%$ 				&$86.29\%$			&$23.99\%$ \\
\bottomrule
\end{tabular}
\end{center}
\caption{
    Evaluation of different methods on the NIH dataset. Our proposed framework achieves state-of-the-art by a large margin compared with previous state-of-the-arts.
}
\label{Tab:NIHC2FSegmentation}
\end{table}

\footnotetext[1]{The results are reported by our runs using the same cross-validation splits where code is available from their GitHub: \url{https://github.com/yulequan/HeartSeg}.}

First of all, our overall coarse-to-fine framework outperforms previous state-of-the-art by nearly $2.2\%$ (Cai \emph{et al.}~\cite{cai2017improving} and Zhou \emph{et al.}~\cite{zhou2017fixed}) in terms of average DSC, which is a large improvement. The lower standard deviation of DSC shows that our method is the most stable and robust across all different CT cases. Although the enhancement of max DSC of our framework is small due to saturation, the improvement of the min DSC over the second best (Dou \emph{et al.}~\cite{dou20173d}) is from $62.53\%$ to $69.62\%$, which is a more than $7\%$ advancement. The worst case almost reaches $70\%$, which is a reasonable and acceptable segmentation result. After coarse-to-fine, the segmentation result of the worst case is improved by more than $11\%$ after the 3D-based refinement from the 3D-based coarse result. The overall average DSC was also improved by $1.41\%$, which proves the effectiveness of our framework.

\begin{figure*}
\centering
	\includegraphics[width=0.85\linewidth]{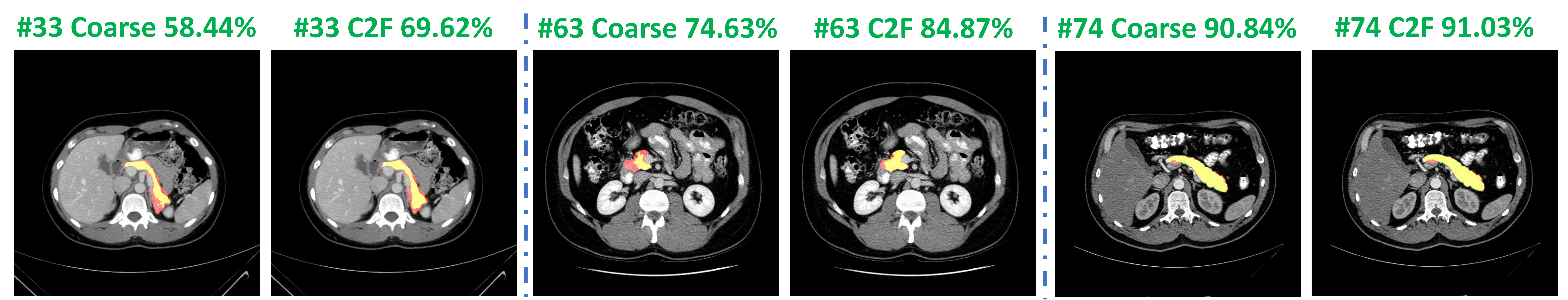}
   \caption{Examples of segmentation results reported by ``ResDSN Coarse" and ``ResDSN C2F" on a same slice in the axial view from NIH case $\#33$, $\#63$ and $\#74$, respectively. Numbers after ``Coarse" or ``C2F" mean testing DSC. Red, green and yellow indicate the ground truth, prediction and overlapped regions, respectively. Best viewed in color.
}
\label{fig:NIHC2F3}
\end{figure*}

% \textcolor{red}{Put some visualization results here by comparisons including failure cases and why}.
As shown in Fig~\ref{fig:NIHC2F3}, we report the segmentation results by ``ResDSN Coarse" and ``ResDSN C2F" on the same slice for comparison. Note that yellow regions are the correctly predicted pancreas. For the NIH case $\#33$, which is the min DSC case reported by both ``ResDSN Coarse" and ``ResDSN C2F", the ``ResDSN C2F" successfully predict more correct pancreas regions at the bottom, which is obviously missed by ``ResDSN Coarse". If the coarse segmentation is bad, \emph{e.g.}, case $\#33$ and $\#63$, our 3D coarse-to-fine can significantly improve the segmentation results by as much as $10\%$ in DSC. However, if the coarse segmentation is already very good, \emph{e.g.}, case $\#74$, our proposed method cannot improve too much. We conclude that our proposed ``ResDSN C2F" shows its advancement over 2D methods by aggregating rich spatial information and is more powerful than other 3D methods on the challenging pancreas segmentation task.

% As shown in Fig~\ref{fig:MinDSCImprovement}, we display three segmentation examples of our ``ResDSN" in comparison with the manual ground truth each case shown in one row. First of all, our ResDSN outputs a 3D segmentation with a very natural and smooth boundary while the manual segmentation is jagged to some extent since the annotation was labeled slice by slice. For the case 2, it seems that our ResDSN has a very large over-estimation at the pancreas head (left bottom region) but we got confirmed from experienced radiologists that our prediction should be correct while the over-estimation is caused by a missing annotation. For the case 7, ResDSN gives an almost perfect segmentation; and for the case 18, the predicted result shows more under-estimation.

% \begin{figure}
% \centering
% 	\includegraphics[width=0.45\textwidth]{figures/ResDSNPrediction.pdf}
%    \caption{An illustration the predicted segmentation of our framework on NIH dataset. In the third column which shows the signed distance map between manual segmentation and ResDSN prediction, green regions mean a perfect prediction while yellow and blue denote under-estimation and over-estimation, respectively. Best viewed in color.
% }
% \label{fig:MinDSCImprovement}
% \end{figure}

%\vspace{-0.28cm}

\subsubsection{JHMI Pathological Cyst Dataset}
\label{sec:JHMICystDataset}
We verified our proposed idea on the JHMI pathological cyst dataset~\cite{zhou2017deep} of abdominal CT scans as well. Different from the NIH pancreas dataset, which only contains healthy pancreas, this dataset includes pathological cysts where some can be or can become cancerous. The pancreatic cancer stage largely influences the morphology of the pancreas~\cite{lasboo2010morphological} that makes this dataset extremely challenging for considering the large variants.

This dataset has a total number of $131$ contrast-enhanced abdominal CT volumes with human-labeled pancreas annotations. The size of CT volumes is $512\times512\times D$, where $D\in [358, 1121]$ that spans a wider variety of thickness than one of the NIH dataset. Following the training protocol~\cite{zhou2017deep}, we conducted $4$-fold cross validation on this dataset where each testing fold has $33, 33, 32$ and $33$ cases, respectively. We trained networks illustrated in Fig.~\ref{fig:ResDSN} in both the $\textit{\textbf{coarse}}$ and $\textit{\textbf{fine}}$ stage with the same training settings as on the NIH except that we trained a total of $300\rm{,}000$ iterations on this pathological dataset since a pancreas with cysts is more difficult to segment than a normal case. In the testing phase, we vote the prediction map of overlapped regions from the sliding window and ignore small isolated pancreas predictions whose portions are smaller than $0.05$ of the total prediction. As shown in Table.~\ref{Tab:JHUC2FSegmentation}, we compare our framework with only one available published results on this dataset. ``ResDSN C2F" achieves an average ${80.56\%}$ DSC that consistently outperforms the 2D based coarse-to-fine method~\cite{zhou2017deep}, which confirms the advantage of leveraging the rich spatial information along three axes. What's more, the ``ResDSN C2F" improves the ``ResDSN Coarse" by ${2.60\%}$ in terms of the mean DSC, which is a remarkable improvement that proves the effectiveness of the proposed 3D coarse-to-fine framework. Both ~\cite{zhou2017deep} and our method have multiple failure cases whose testing DSC are $0$, which indicates the segmentation of pathological organs is a more tough task. Due to these failure cases, we observe a large deviation on this pathological pancreas dataset compared with results on the NIH healthy pancreas dataset.

 \begin{table}[htb]
 \footnotesize
 \begin{center}
 \begin{tabular}{lc l c}\toprule
 Method           & {Mean DSC} \\
 \hline
 %{ResDSN C2F (Ours)}  		& $\bm{80.17\pm{13.63}}\%$ \\
 %{ResDSN Coarse (Ours)}  		& $77.32\pm{13.59}\%$ \\
 {ResDSN C2F (Ours)}  		& $\bm{80.56\pm{13.36}}\%$ \\
 {ResDSN Coarse (Ours)}  		& $77.96\pm{13.36}\%$ \\
 \hline
 {Zhou \emph{et al.}~\cite{zhou2017deep}} 	& $79.23\pm{9.72}\%$\\
 \bottomrule
 \end{tabular}
 \end{center}
 \caption{
     Evaluations on the JHMI pathological pancreas.
 }
 \label{Tab:JHUC2FSegmentation}
 \vspace{-0.2in}
 \end{table}
 
 \subsubsection{JHMI Pancreas Dataset}
 \label{sec:JHMIPancreas}
In order to further validate the superiority of our 3D model, We also evaluate our approach on a large high-quality dataset collected by the radiologists in our team.
This dataset contains $305$ contrast-enhanced abdominal CT volumes, and each of them is manually labeled with pancreas masks. Each CT volume consists of $319 \sim 1051$ slices of $512 \times 512$ pixels, and have voxel spatial resolution of $([0.523 \sim 0.977] \times [0.523 \sim 0.977]\times 0.5)\textup{mm}^3$. Following the training protocol~\cite{roth2015deeporgan}, we perform $4$-fold cross-validation in a random split from all patients for training and testing folds, where each testing fold has $77, 76, 76$ and $76$ cases, respectively. We demonstrate the superiority of our 3D model\footnote{The coarse model is used for comparison since it is the basis of our framework} by comparing with the 2D baseline~\cite{zhou2017fixed} (see Table~\ref{Tab:JHUC2FPancreas}).

 \begin{table}[htb]
 \footnotesize
 \begin{center}
 \begin{tabular}{lc l c}\toprule
 Method           & {Mean DSC}  & {Max DSC}  & {Min DSC} \\
 \hline
 %{ResDSN C2F (Ours)}  		& $\bm{80.17\pm{13.63}}\%$ \\
 %{ResDSN Coarse (Ours)}  		& $77.32\pm{13.59}\%$ \\
%  {ResDSN C2F (Ours)}  		& $\bm{80.56\pm{13.36}}\%$ \\
 {ResDSN Coarse (Ours)}  		& $\bm{87.84\pm{7.27}}\%$  & $\bm{95.27}\%$ & $0.07\%$ \\
%  \hline
 {Zhou \emph{et al.}~\cite{zhou2017deep}} 	& $84.99\pm{7.42}\%$ & $93.45\%$ & $\bm{3.76}\%$\\
 \bottomrule
 \end{tabular}
 \end{center}
 \caption{
     Evaluations on the JHMI pancreas dataset.
 }
 \label{Tab:JHUC2FPancreas}
 \vspace{-0.2in}
 \end{table}

\subsubsection{Ablation Study}\label{sec:ablation}
% under-estimation versus over-estimation
In this section, we conduct the ablation studies about residual connection, time efficiency and deep supervision to further investigate the effectiveness and efficiency of our proposed framework for pancreas segmentation.

\paragraph{\textbf{Residual Connection}}
We discuss how different combinations of residual connections contribute to the pancreas segmentation task on the NIH dataset. All the residual connections are implemented in the element-wise sum and they shared exactly the same deep supervision connections, cross-validation splits, data input, training and testing settings except that the residual structure is different from each other. As given in Table.~\ref{Tab:NIHResSegmentation}, we compare four configurations of residual connections of 3D based networks only in the $\textit{\textbf{coarse}}$ stage. The major differences between our backbone network ``ResDSN" with respect to ``FResDSN", ``SResDSN" and ``DSN" are depicted in Table.~\ref{Tab:Config3DSegmentation}. ``ResDSN" outperforms other network architectures in terms of average DSC and a small standard deviation even through the network is not as sophisticated as ``FResDSN", which is the reason we adopt ``ResDSN" for efficiency concerns in the \textit{\textbf{coarse}} stage.

\begin{table}[tb]
\footnotesize
\begin{center}
\begin{tabular}{lccc}\toprule
Method           & {Mean DSC}       		& {Max DSC}    		& {Min DSC} \\
\hline
{ResDSN Coarse (Ours)}  		& $\bm{83.18\pm{6.02}}\%$ 				& $91.33\%$       	&$58.44\%$ \\
\hline
{FResDSN Coarse} 	& $83.11\pm{6.53}\%$ 				&$91.34\%$ 			& $61.97\%$\\
{SResDSN Coarse} &$82.82\pm{5.97}\%$ 				&$90.33\%$			&$62.43\%$ \\
{DSN~\cite{dou20173d} Coarse}		&$82.25\pm{5.91}\%$ 				&$90.32\%$			&$62.53\%$ \\
\bottomrule
\end{tabular}
\end{center}
\caption{
    Evaluation of different residual connections on NIH.
}
\label{Tab:NIHResSegmentation}
\end{table}

\paragraph{\textbf{Time Efficiency}}
We discuss the time efficiency of the proposed coarse-to-fine framework with a smaller overlap in the \textit{\textbf{coarse}} stage for the low consuming time concern while a larger one in the \textit{\textbf{fine}} stage for the high prediction accuracy concern. The overlap size depends on how large we choose $n$ defined in Sec~\ref{Sec:NetworkTrTs}. We choose $n=6$ during the coarse stage while $n=12$ during the fine stage. Experimental results are shown in Table~\ref{Tab:NIHTimeCost}. ``ResDSN Coarse" is the most efficient while the accuracy is the worst among three methods, which makes sense that we care more of the efficiency to obtain a rough pancreas segmentation. ``ResDSN Fine" is to use a large overlap on an entire CT scan to do the segmentation which is the most time-consuming. In our coarse-to-fine framework, we combine the two advantages together to propose ``ResDSN C2F" which can achieve the best segmentation results while the average testing time cost for each case is reduced by $36\%$ from $382$s to $245$s compared with ``ResDSN Fine". In comparison, it takes an experienced board certified Abdominal Radiologist 20 mins for one case, which verifies the clinical use of our framework.

\begin{table}[tb]
\footnotesize
\begin{center}
\begin{tabular}{lccc}\toprule
Method				& Mean DSC			& $n$ &Testing Time (s) \\
% $111$ + $134$
\hline
{ResDSN C2F (Ours)}  & $\bm{84.59\pm{4.86}}\%$ & $6\&12$	& $245$\\
{ResDSN Coarse (Ours)}  & $83.18\pm{6.02}\%$ & $6$	& $\bm{111}$\\
{ResDSN Fine (Ours)}	& $83.96\pm{5.65}\%$ & $12$	& $382$\\
\bottomrule
\end{tabular}
\end{center}
\caption{
    Average time cost in the testing phase, where $n$ controls the overlap size of sliding windows during the inference.
}
% \vspace{-0.20in}
\vspace{-0.4cm}
\label{Tab:NIHTimeCost}
\end{table}

\paragraph{\textbf{Deep Supervision}}
We discuss how effective of the auxiliary losses to demonstrate the impact of the deep supervision on our 3D coarse-to-fine framework. Basically, we train our mainstream networks without any auxiliary losses for both coarse and fine stages, denoted as ``Res C2F", while keeping all other settings as the same, \emph{e.g.}, cross-validation splits, data pre-processing and post-processing. As shown in Table~\ref{Tab:DeepSupervision}, ``ResDSN C2F" outperforms ``Res C2F" by $17.79\%$ to a large extent on min DSC and $0.53\%$ better on average DSC though it's a little bit worse on max DSC. We conclude that 3D coarse-to-fine with deep supervisions performs better and especially more stable on the pancreas segmentation.

\begin{table}[tb]
\footnotesize
\begin{center}
\begin{tabular}{lccc}\toprule
Method           & {Mean DSC}       		& {Max DSC}    		& {Min DSC} \\
\hline
{ResDSN C2F (Ours)}  		& $\bm{84.59\pm{4.86}}\%$ 				& $91.45\%$       	&$\bm{69.62}\%$ \\
%{ResDSN Coarse (Ours)}  		& $83.18\pm{6.02}\%$ 				& $91.33\%$       	&$58.44\%$ \\
{Res C2F} &$84.06\pm{6.51}\%$ 				&$\bm{91.72}\%$			&$51.83\%$ \\
%{Res Coarse}		&$83.38\pm{5.48}\%$ 				&$91.45\%$			&$63.75\%$ \\
\bottomrule
\end{tabular}
\end{center}
\caption{
    Ablation study of the deep supervision on NIH.
}
\vspace{-0.4cm}
\label{Tab:DeepSupervision}
\end{table}

\vspace{-0.3cm}

\subsection{Adversarial Attack and Defense}
In spite of the success of 3D learning models such as our proposed ResDSN, the robustness of neural networks for clinical applications remains a concern. In this section, we first show that our well-trained 3D model can be easily led to failure under imperceptible adversarial perturbations (see Sec.~\ref{sec:AttackExp}), and then investigate how to improve the adversarial robustness by employing adversarial training (see Sec.~\ref{sec:DefendExp}). We evaluate our approach by performing standard $4$-fold cross-validation on the JHMI pancreas dataset since this dataset is the largest in scale and has the best quality (see Sec.~\ref{sec:JHMIPancreas}).

\subsubsection{Robustness Evaluation}
\label{sec:AttackExp}

To evaluate the robustness of our well-trained 3D model, we attack the ResDSN Coarse model following the methods in Sec.~\ref{sec:AttackMethods}.  
For both attacking methods, \emph{i.e.}, FGSM and I-FGSM, we set $\epsilon = 0.03\Lambda$ so that the maximum perturbation can be small enough compared with the range of the truncated intensity value ($\Lambda$)\footnote{Since the raw intensity values are to be in $[-100, 240]$ during preprocessing (see Sec.~\ref{Sec:NetworkTrTs}), here we set $\Lambda = 240 - (-100) = 340$ accordingly.}. Specially in the case of I-FGSM, the total iteration number $N$ and the step size $\alpha$ are set to be $5$ and $0.01\Lambda$, respectively. Following the test strategy in the coarse stage, we first compute the loss gradients of the $64 \times 64 \times 64$ sub-volumes\footnote{For implementation simplicity and efficiency, we ignored the sub-volumes only containing the background class when generating adversarial examples.} obtained by a sliding window policy, %$\nabla_{\mathbf{X_{sub}}} \mathcal{L}(\mathbf{X_{sub}}; \bm{\Theta})$
 and these gradients are then combined to calculate the final loss gradient map $\nabla_{\mathbf{X}} \mathcal{L}(\mathbf{X}; \bm{\Theta})$ of each whole CT volume. The combine approach is also similar as the testing method described in Sec.~\ref{sec:JHMICystDataset}, \emph{i.e.}, taking the average of loss gradient if a voxel is in the overlapped region. According to Eq.~\ref{eq:FGSM} and Eq.~\ref{eq:IFGSM update}, the overall loss gradient can be used to generate adversarial examples which can then attack the 3D model for the purpose of robustness evaluation.

\begin{table}[tb]
\footnotesize
\begin{center}
\begin{tabular}{lcccc}\toprule
Attack Methods           & {Clean}       		& {FGSM~\cite{goodfellow2014explaining}}    		& {I-FGSM~\cite{kurakin2016scale}}  &{Drop}\\
\hline
{ResDSN Coarse}  		& $\bm{87.84\pm{7.27}}\%$				& $42.68\%$       	&$2.01\%$  &$85.83\%$\\
{Adversarially-trained ResDSN Coarse}  		& $79.09\pm{12.10}\%$ 				& $\bm{67.58}\%$       	&$\bm{65.98}\%$  &$\bm{13.11}\%$\\

\bottomrule
\end{tabular}
\end{center}
\caption{
    Comparative evaluation of the 3D segmentation model on clean (indicated by ``Clean'' in the table) and adversarial examples. Different attack methods,\emph{i.e.}, FGSM and I-FGSM, are used for generating the adversarial examples. We report the average accuracy and Dice overlap score along with the $\%$ maximum drop in performance on adversarial examples with respect to performance on clean data.  
}
\label{Tab:advResDSN}
\end{table}

\begin{figure*}
 	\centering
 	\includegraphics[width=1\columnwidth]{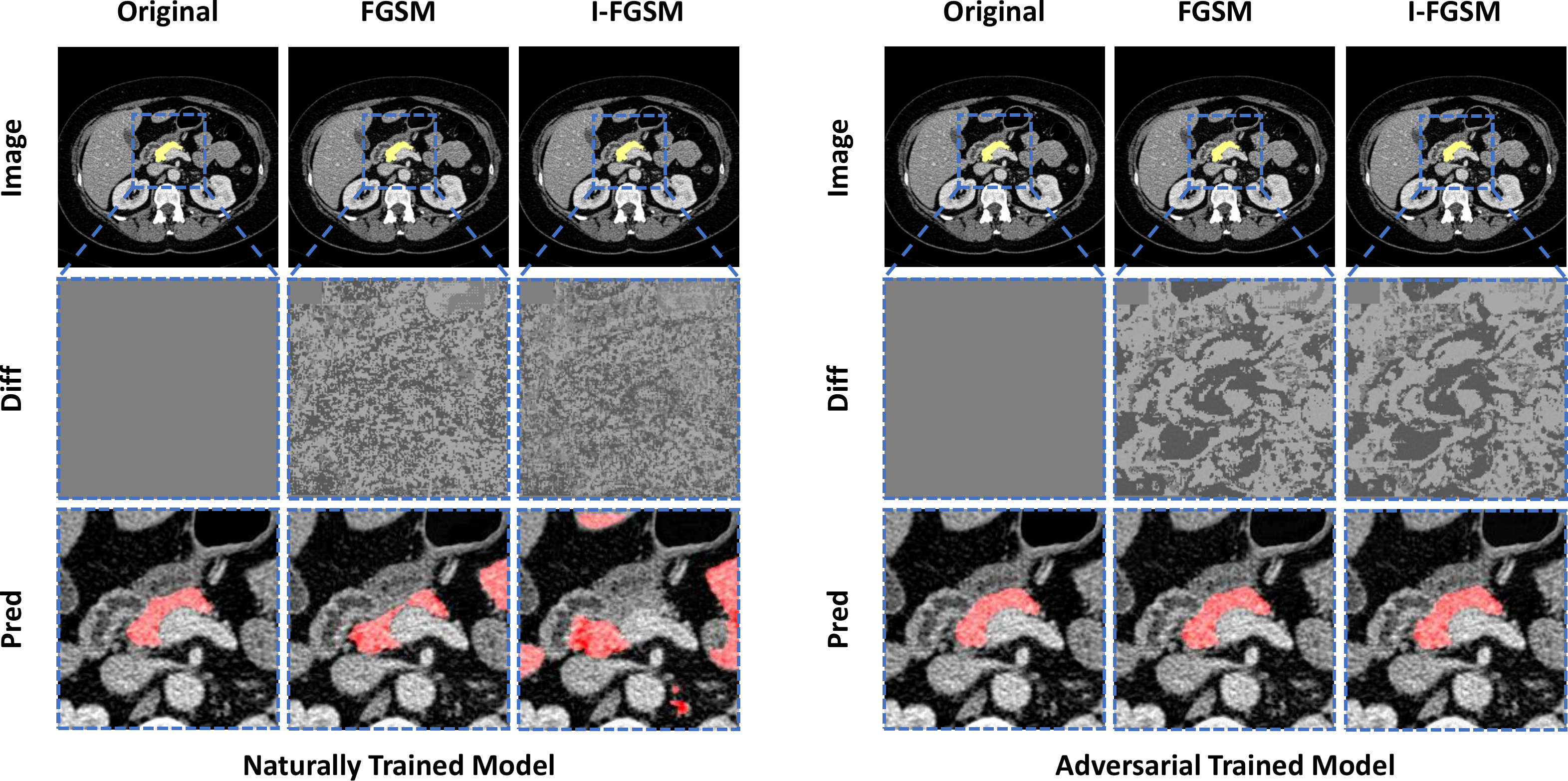}
 	\caption{Qualitative comparison of adversarial examples and their effects on model predictions.
    Note that the added perturbation is effectively imperceptible to the human eye, and the difference between the original image and the adversarial image has been magnified by 5x(values shifted by 128) for a better visualization. Contrasting with prediction on original images, the crafted examples are able to successfully fool the models into  generating incorrect segmentation maps. Meanwhile adversarial training can effectively alleviate such negative influence of adversarial attacks, hence improving the performance to a reasonable level. Image differences and predictions are zoomed in from the axial view to better visualize the finer details.}
 	\label{fig:qualitative}
 	\vspace{-1em}
\end{figure*}

\subsubsection{Defending Against Adversarial Attacks}
\label{sec:DefendExp}
To improve the adversarial robustness of our 3D segmentation model, we apply the adversarial training policy as described in Sec.~\ref{sec: defense}. During each training iteration, $\mathbf{X_{adv}}$ is first randomly sampled in the $\epsilon\text{-ball}$ and then updated by I-FGSM so that $\mathcal{L}(\mathbf{X_{adv}};\bm{\Theta})$ can be maximized. 
%to get $\mathbf{X_{adv}}$ in the $\epsilon\text{-ball}$ of an original sub-volume which is difficult enough for current deep model 
%(\emph{i.e},~$\max_{\mathbf{r}\in \mathcal{S}} \mathcal{L}(\mathbf{X}+\mathbf{r};\bm{\Theta})$). 
Afterwards $\mathbf{X_{adv}}$ is fed to the model instead of $\mathbf{X}$ to update the parameter $\mathbf{\Theta}$. Note that we set the same maximum perturbation $\epsilon$, iteration number $N$ and step size $\alpha$ as in Sec.~\ref{sec:AttackExp}. Similar to the training process described in Sec.~\ref{sec:NIHDataset}, our model is trained by SGD optimizer with a $128$ mini-batch, a $0.9$ momentum, a base learning rate to be $0.08$ via polynomial decay (the power is $0.9$) in a total of $10,000$ iterations, and the weight decay $0.0005$.

\subsubsection{Results and Discussion}
All attack and defense results are summarized in Table~\ref{Tab:advResDSN}. We can see that both attack methods, \emph{i.e.} FGSM and I-FGSM, can successfully fool the well-trained 3D ResDSN into producing incorrect prediction maps. More specifically, the dramatic performance drop of I-FGSM, \emph{i.e.},$85.83\%$ (from $87.84\%$ to $2.01\%$), suggests low adversarial robustness of the original model. Meanwhile the maximum performance drop decreases from $85.83\%$ to $13.11\%$, indicating that our adversarially-trained model can largely alleviate the adversarial effect and hence improving the robustness of our 3D model. Note that our baseline with ``Clean'' training has $87.84\%$ accuracy when tested on clean images, whereas its counterpart with adversarial training obtains $79.09\%$. This tradeoff between adversarial and clean training has been previously observed in~\cite{tsipras2018robustness}. We hope this tradeoff can be better studied in future research.

We also show a qualitative example in Fig.~\ref{fig:qualitative}. As can be observed from the illustration, adversarial attacks to naturally trained 3D ResDSN induces many false positives, which makes the corresponding outcomes noisy. On the contrary, the adversarially trained 3D model yields similar performances even after applying I-FGSM. More specifically, the original average Dice score of 3D ResDSN is $89.30\%$, and after applying adversarial attack the performance drops to $48.45\%$ and $6.06\%$ with FGSM and I-FGSM respectively. However, when applying the same attack methods to the adversarial trained model, the performance only drops from $86.41\%$ to $80.32\%$ and $79.56\%$ respectively. In other words, employing adversarial training decreases the performance drop from $83.24\%$ to only $6.85\%$. This promising result clearly indicates that our adverarially-trained model can largely improve the adversarial robustness.
%For the original image, the performance of our adversarial trained model is $86.41\%$.

\vspace{-0.5cm}
\section{Conclusion}
\vspace{-0.5cm}
In this chapter, we proposed a novel 3D network called ``ResDSN" integrated with a coarse-to-fine framework to simultaneously achieve high segmentation accuracy and low time cost. The backbone network ``ResDSN" is carefully designed to only have long residual connections for efficient inference. In addition, we also analyzed the threat of adversarial attacks on our framework and showed how to improve the robustness against the attack. 
%For the first time in the field of medical image analysis, 
Experimental evidence indicates that our adversarially trained model can largely improve adversarial robustness than naturally trained ones. 

To our best knowledge, the proposed 3D coarse-to-fine framework is one of the first works to segment the challenging pancreas using 3D networks which leverage the rich spatial information to achieve the state-of-the-art. We can naturally apply the proposed idea to other small organs, \emph{e.g.}, spleen, duodenum and gallbladder, \emph{etc},  In the future, we will target on error causes that lead to inaccurate segmentation to make our framework more stable, and extend our 3D coarse-to-fine framework to cyst segmentation which can cause cancerous tumors, and the very important tumor segmentation~\cite{zhu2018multi} task.\\

%On widely-used datasets, the worst segmentation case is experimentally improved a lot by our coarse-to-fine framework. What's more, our coarse-to-fine framework can work on both normal and abnormal pancreases to achieve good segmentation accuracy.
% and can be well generalized to segment other challenging organs as well.

%Though this work mainly focuses on segmentation for the pancreas, we can naturally apply the proposed idea to other small organs, \emph{e.g.}, spleen, duodenum and gallbladder, \emph{etc},  In the future, we will target on error causes that lead to inaccurate segmentation to make our framework more stable, and extend our 3D coarse-to-fine framework to cyst segmentation which can cause cancerous tumors, and the very important tumor segmentation~\cite{zhu2018multi} task.\\

{\small
\bibliographystyle{splncs03}
\bibliography{egbib}
}
\end{document}